%% file: Main.tex
\documentclass{article}
\usepackage[preprint]{spconf}
\copyrightnotice{\copyright\ all rights reserved}
\toappear{Preprint}
\usepackage{amsmath,graphicx}
\usepackage{amssymb}
\usepackage{subcaption}

\usepackage{algorithm}
\usepackage{algpseudocode}
\usepackage[table,xcdraw]{xcolor}

\title{Large-scale kernelized Granger causality to infer topology of directed graphs with applications to brain networks}
\name{M. Ali Vosoughi$^{\star}$ and Axel Wismüller$^{\star, \dagger, \ddagger, \mathsection}$\thanks{This research was funded by Ernest J. Del Monte Institute for Neuroscience Award from the Harry T. Mangurian Jr. Foundation. This work was conducted as a Practice Quality Improvement (PQI) project related to American Board of Radiology (ABR) Maintenance of Certificate (MOC) for Prof. Dr. Axel Wismüller. This work is not being and has not been submitted for publication or presentation elsewhere.}}
\address{$^{\star}$ Department of Electrical and Computer Engineering, University of Rochester, NY, USA\\
 $^{\dagger}$ Department of Imaging Sciences, University of Rochester, NY, USA\\
$^{\ddagger}$ Department of Biomedical Engineering, University of Rochester, NY, USA\\
$^{\mathsection}$ Faculty of Institute of Clinical Radiology, Ludwig Maximilian University,
Munich, Germany}
\begin{document}
%\ninept
%
\maketitle
\begin{abstract}
Graph topology inference of network processes with co-evolving and interacting time-series is crucial for network studies. Vector autoregressive models (VAR) are popular approaches for topology inference of directed graphs; however, in large networks with short time-series, topology estimation becomes ill-posed. The present paper proposes a novel nonlinearity-preserving topology inference method for directed networks with co-evolving nodal processes that solves the ill-posedness problem. The proposed method, large-scale kernelized Granger causality (lsKGC), uses kernel functions to transform data into a low-dimensional feature space and solves the autoregressive problem in the feature space, then finds the pre-images in the input space to infer the topology. Extensive simulations on synthetic datasets with nonlinear and linear dependencies and known ground-truth demonstrate significant improvement in the Area Under the receiver operating characteristic Curve ( AUC )  of the receiver operating characteristic for network recovery compared to existing methods. Furthermore, tests on real datasets from a functional magnetic resonance imaging (fMRI) study demonstrate 96.3 percent accuracy in diagnosis tasks of schizophrenia patients, which is the highest in the literature with only brain time-series information.
\end{abstract}

\input{intro}

\vspace{-0.15in}
\section{PRELIMINARIES}
Consider an $N$-node directed graph, whose topology is unknown, but time-series $\{y_{it}\}^T_{t=1}$ are observed per node $i$, over $T$ time intervals. Linear VAR model states that each $y_{jt}$ is a linear combination of the time-lagged versions of the measurements $\{\{y_{i(t-\ell )}\}^N_{i=1}\}^L_{\ell=1}$; see \textit{e.g.}, \cite{chen2011vector, dsouza2017exploring}
\begin{equation} 
y_{jt}=\sum_{i=1}^{N}\sum_{\ell=1}^{L}a_{ij}^{\ell}y_{j(t-\ell)}+e_{jt}, \label{eqn:linear_autoregressive_model}
\end{equation}
with $a_{ij}^{\ell}$ as model coefficients encoding the causal structure of the network over a lag of $\ell$ time points. Based on the error covariance matrix $\Sigma$ of the full model and the error covariance matrix $\Sigma^{i-}$ of the model without $\{y_{it}\}^T_{t=1}$, the degree of information flow from node $i$ to node $j$ can be quantified by $\delta_{ij}=\ln({\Sigma_{j}^{i-}}/{\Sigma_{j}})$, 
where $\Sigma_{j}^{i-}$ and $\Sigma_{j}$ denote the diagonal entries of $\Sigma^{i-}$ and $\Sigma$ associated to $\{y_{jt}\}^T_{t=1}$, respectively. Within this framework, a Granger Causality Index (GCI) \cite{dsouza2017exploring} can be defined in terms of linear predictability \cite{granger1988some}, that is, a causal link exists between nodes $i$ and $j$ only if there exist $\delta_{ij}\neq 0$. Consequently, the topology of the graph directly will be obtained by adjacency matrix $[\mathbf{\Delta}]_{ij}=\delta_{ij}$.

Let $\mathbf{A}^{\ell}\in\mathbb{R}^{N\times N}$ denote the “time-lagged” VAR model parameters matrix, with $[\mathbf{A}^{\ell}]_{ij}=a_{ij}^{\ell}$. Given the multivariate time-series $\{\mathbf{y}_t\}^T_{t=1}$, where $\mathbf{y}_{t}:=[{y}_{1t},\ldots,{y}_{Nt}]^\top$, the goal is to estimate the model parameter matrices $\{\mathbf{A}^{\ell}\}_{\ell=1}^L$, and consequently unveil the hidden network topology. Although generally known, one can readily deduce $L$ via standard model order selection tools, \textit{e.g.}, the Bayesian information criterion (BIC)\cite{chen1998speaker}, or Akaike's Information Criterion (AIC) \cite{bozdogan1987model}.

In a well-defined system with known topology, several approaches have been proposed to learn the model parameters; see \textit{e.g.}, \cite{chen2011vector,roebroeck2005mapping}. However, the model's complexity increases by increasing $N$ for the same $T$ in the system since redundant variables can lead to ill-posed problems \cite{angelini2010redundant, dsouza2017exploring}.

Accordingly, a method that estimates multivariate interactions while reducing redundancy would be desired. In \cite{dsouza2017exploring}, a VAR model is proposed to learn linear causal influences for large-scale systems by reducing redundancy through projecting data to principal space. Albeit conceptually simple and computationally tractable, the linear VAR of \cite{dsouza2017exploring} is incapable of capturing nonlinear dependencies inherent to complex networks, \textit{e.g.}, the human brain. The present paper proposes a nonlinear VAR model by reducing the redundancies in \eqref{eqn:linear_autoregressive_model} using a nonlinear transformation of data to the feature space.
\newline
\textbf{Problem Statement.}  Given nodal measurements $\{y_{it}\}^T_{t=1}$ in \eqref{eqn:linear_autoregressive_model}, the goal is to estimate $\{\hat{y}_{it}\}^T_{t=L+1}$ in a low-dimensional feature space, encompassing nonlinear dependencies between time-series, to bypass estimation of parameter matrix $\{\mathbf{A}^{\ell}\}_{\ell=1}^L$ in the input space, subsequently transform the predicted $\{\hat{y}_{it}\}^T_{t=L+1}$ to the input space to obtain the unknown adjacency matrix $\mathbf{\Delta}$ of causal relationships.

\vspace{-0.1in}
\section{Large-scale Kernelized Granger Causality}
Time-series $\{y_{jt}\}_{t=1}^T$ for $j=1,\ldots,N$ are series with linear or nonlinear causal dependencies between $N$ nodes. 
The aim is to transfer the equation \eqref{eqn:linear_autoregressive_model} to the space of the principal components with lower dimensions, encompassing nonlinear dependencies. 
We define function $\phi(.)$ as follows
\begin{equation}\label{eqn:kernel_transform_mapping_phi} 
    \phi({y}_{jt}): \mathcal{Y}\rightarrow \mathcal{H}~~, ~~{y}_{jt}\mapsto \phi({y}_{jt}),
\end{equation}
to transfer data from the input space to the feature space.
A feature space related to the principal components with the maximum explained variance of data would be desired to solve the autoregressive problem, encompassing nonlinear dependencies.
Defining $P<\infty$ as the dimensions of the feature space, $\{\phi_{p}(.)\}^P_{p=1}$ as the basis of feature space with (possibly) known kernel function, where $\{\phi_{p}(.)\}^P_{p=1}$ spans a finite subspace of the Hilbert space related to finite input space, and $\mathbf{\phi}(y_{it}) :=[\phi_{1}(y_{it}), \ldots, \phi_{P}(y_{it})]^\top$. A projection of input data ${y}_{jt}$ onto the $p$ -th principal component  would be desired, such  that the major proportion of data variation is explained by a few principal components; see \cite{scholkopf2002learning}. 

Defining $\mathbf{\Phi}_j:=[\mathbf{\phi}(y_{j1}),\ldots,\mathbf{\phi}(y_{jT})]^\top$, the covaraince matrix of the data in $\mathcal{H}$ will be $\mathbf{C} = \frac{1}{N}\sum^N_{j=1}{\mathbf{\Phi}_j}^\top\mathbf{\Phi}_j$.
To find $\{\phi_{p}(.)\}^P_{p=1}$ as the basis of feature space, eigenvalues $\lambda_p\ge 0: \{\lambda_{1}\ge \ldots\ge \lambda_{P}\}$ would be desired. Consequently, in the feature space  $\exists ~ \mathbf{v}_p\in\mathcal{H}\setminus\{0\}$ such that $\lambda_p\mathbf{v}_p=\mathbf{C}\mathbf{v}_p$, and $<\mathbf{v}_p,\mathbf{v}_p>=1$ to ensure orthonormality of eigenvectors $\mathbf{\{v_p}\}_{p=1}^P$ in the feature space. Consequently, exists $\exists ~ \{\mathbf{\beta_{j}}\}_{j=1}^N : \mathbf{\beta_{j}}\in \mathbb{R}^P$ , that eigenvectors can be written as a linear combination for features $\mathbf{v}_p = \sum^N_{j=1}\mathbf{\beta_{j}}{\mathbf{\Phi}_j}^\top$.
Acknowledging that $\mathbf{\phi}(.)$ is probably a known function, \textit{e.g.} radial basis function (RBF), obtaining the eigenvectors is equivalent to estimating the coefficients $\mathbf{\beta_{j}}$.
Thus, any function $\phi(.)$ (equivalently $\{\phi_{p}^{\ell}(.)\}^P_{p=1}$) can be written as a linear combination of $\{\mathbf{v}_p\}_{p=1}^P$ to project data onto eigenvectors with the highest explained variance in the feature space.
Using the transformation of $\{\phi_{p}^{\ell}(.)\}^P_{p=1}$, the feature space representation of the autoregressive equation \eqref{eqn:linear_autoregressive_model} can be written as \cite{kallas2013kernel}
\begin{equation} 
\eta_{pt}=\sum_{\ell=1}^{L}\sum_{j=1}^{N}\alpha_{jp}^{\ell}\phi_{p}({y}_{j(t-\ell)})+\varepsilon_{pt} \label{eqn:linear_autoregressive_feature_space}
\end{equation}
where $\{\eta_{pt}\}^T_{t=1}$ for $p=1,\ldots,P$ is the transformed time-series into the feature space, and $\{\varepsilon_{pt}\}^P_{p=1}$ is the  error at time slot $t$ in the feature space.
One can use models such as linear regression, LS with $\ell 1$-regularization (the lasso), $\ell 2$-regularization (ridge regression), or elastic-net regression to minimize error $\{\eta_{pt}\}^P_{p=1}$ and obtain the unknown coefficients $\{\alpha_{jp}^{\ell}\}$; see \cite{friedman2010regularization}.
Once the $\{\eta_{pt}\}^P_{p=1}$ in \eqref{eqn:linear_autoregressive_feature_space} are obtained , the predictions $\{\hat{\eta}_{pt}\}^P_{p=1}$ will be transformed into the input space to infer the graph topology.
To transform $\{\hat{\eta}_{pt}\}^P_{p=1}$ into the input space, the pre-image ${\gamma}(\hat{\eta}_{pt}):\mathcal{H}\rightarrow\mathcal{Y},~~\hat{\eta}_{pt}\mapsto{\gamma}(\hat{\eta}_{pt})$, would require the estimation \cite{scholkopf2002learning}
\begin{equation}
    \hat{y}_{jt} = \arg \min \limits _{\hat{y}_{jt}\in \mathcal{Y}} \Vert{y}_{jt}-\gamma(\phi(\hat{y}_{jt}))\Vert^2,
\end{equation} 
which is a nonlinear optimization, lacking the computational efficiency, applicability to pre-images with discrete variables, and a guaranteed optimum solution in general; see \cite{scholkopf2002learning, bakir2003learning}. Interestingly, both the inputs $\{y_{it}\}^T_{t=1}$ and their corresponding mappings are present for $\phi({y}_{jt}): \mathcal{Y}\rightarrow \mathcal{H}$.
Defining $\Tilde{\mathbf{y}}_j:=[y_{j1},\ldots,y_{jT}]^\top$, $\mathbf{Y}:=[\Tilde{\mathbf{y}}_1,\ldots,\Tilde{\mathbf{y}}_N]$, $\mathbf{\gamma}(\eta_{pt}) :=[\gamma_{1}(\eta_{pt}), \ldots, \gamma_{N}(\eta_{pt})]^\top$, $\mathbf{\Gamma}_p := [\gamma(\eta_{p1}),$ $ \ldots, $ $\gamma(\eta_{pT})]^\top$,
$\mathbf{\Gamma}:=[\mathbf{\Gamma}_1,\ldots,\mathbf{\Gamma}_P]$, and $\mathbf{\Phi}:=[\mathbf{\Phi}_1,\ldots,\mathbf{\Phi}_N]$, and using the learning problem \cite{bakir2003learning,kwok2002finding}
\begin{equation}
    \mathcal{L} = \arg \min \limits _{\mathbf{\Gamma}}  \Vert\mathbf{Y}- \mathbf{\Gamma}\mathbf{\Phi}\Vert^2,
\end{equation}
one can obtain the coefficients  $\gamma_{jp}:=[\mathbf{\Gamma}]_{jp}$, and consequently the predictions $\hat{y}_{jt}=\sum^P_{p=1}\gamma_{jp} \hat{\eta}_{pt}$ will be obtained; see \cite{bakir2003learning,kwok2002finding}. Simultaneously, the adjacency matrix $\mathbf{\Delta}$ will be inferred in the input space by the corresponding MSE of the error covariance.
The complete algorithm is shown in Algorithm. \ref{alg:complete_alg_kernel_lsGC}, and the notations $\mathbf{Y^{j-}}:=[\Tilde{\mathbf{y}}_1,\ldots,\Tilde{\mathbf{y}}_{j-1},\Tilde{\mathbf{y}}_{j+1},\ldots,\Tilde{\mathbf{y}}_N]$, $\Tilde{\mathbf{\eta}}_{p} :=[\eta_{p1},\ldots,\eta_{pT}]^\top$,  $\mathbf{H}:=[\Tilde{\mathbf{\eta}}_{1},\ldots,\Tilde{\mathbf{\eta}}_{P}]$  are defined for the sake of simplicity. 
\begin{algorithm}
    \caption{Large-scale kernelized Granger causality} \label{alg:complete_alg_kernel_lsGC}
    \begin{algorithmic}
        \Require $L,P,\mathbf{Y},\mathbf{\phi}(.)$
        \Ensure $\mathbf{\Delta}=\{[\mathbf{\Delta}]_i\}_{i=1}^N $
        \State  $\mathbf{\Gamma} \leftarrow  \mathcal{L}  \leftarrow \mathbf{Y} \leftarrow \text{normalize}(\mathbf{Y})$
        \For {$i=1$ to $N$} 
            \State $\mathbf{H}\xleftarrow[]{\mathbf{\Phi}} \mathbf{Y}$
            \For{$\ell=1$ to $L$} 
            $\hat{\mathbf{Y}} \xleftarrow[]{\mathbf{\Gamma}} \hat{\mathbf{H}}$ 
            \EndFor
        \EndFor

        \For {$j=1$ to $N$} 
            \State $\mathbf{H}^{j-}\xleftarrow[]{\mathbf{\Phi}} \mathbf{Y}^{j-}$
            \For {$\ell=1$ to $L$} $\hat{\mathbf{Y}}^{j-} \xleftarrow[]{\mathbf{\Gamma}} \hat{\mathbf{H}}^{j-}$ \EndFor
        \EndFor
        \State $[\mathbf{\Delta}]_{i:}\leftarrow \{\ln(\frac{\mathbf{\Sigma}_{j}^{i-}}{\mathbf{\Sigma}_{j}})\}_{j=1}^N
        \leftarrow \ln(\frac{diag\{cov(\mathbf{Y}^{i-}-\hat{\mathbf{Y}}^{i-})\}}{diag\{cov(\mathbf{Y}-\hat{\mathbf{Y}})\}})$

    \end{algorithmic}

\end{algorithm}

\input{simulations}

\vspace{-0.1in}
\section{Conclusions}
This paper put forth a novel method for the topology inference in large-scale networks and uncovers nonlinear interactions between the nodal processes. By leveraging the feature space,  the proposed method preserves the nonlinearities inherent to complex networks by transforming data into their principal components in the feature space. Further, the pre-images were used to infer the directional network topology in the original space by adopting the learning from mappings between the input and feature spaces. Tests on the synthetic datasets with nonlinear and linear processes and known underlying relationships demonstrate significant improvement in the topology inference of networks. Moreover, tests on the real data from an fMRI study demonstrated 96.3 percent accuracy in the classification task, which is the highest in the schizophrenia diagnosis literature with only brain time-series information. Future directions include broadening the scope of the approach to low-dimensional sparse representations and dynamic networks suitable to real data.

\bibliographystyle{IEEEbib}
\newpage
\ninept
\bibliography{strings,refs}

\end{document}

%% file: intro.tex
\section{INTRODUCTION}
Graph topology inference is crucial in network studies, especially emphasized by recent surges in graph signal processing. Understanding the underlying interactions of network processes shed light on knowledge discovery in multiple domains, \textit{e.g.}, chemistry and medicine, genealogy, climatology, and neuroscience, to name a few. 
Representing the influences beneath a complex network has diverse applications. Nevertheless, to infer the graph topology,  the edges are not directly observable, restricting the nodal observations to the sole source of data acquisition. For instance, in the human brain, time-series data may be recorded from distinct regions of interest (ROI) using resting-state functional magnetic resonance imaging (rs-fMRI); however, the interactions among the recorded time-series are not measurable. 

Vector autoregressive models (VAR) or Granger causality \cite{granger1988some} are widely adopted to infer directional dependencies among the nodal processes. 
Various methods are proposed to infer the directional topology of graphs; see \cite{giannakis2018topology,marques2020signal}. Lasso - Granger learns the causal dependencies based on the variable selection using the Lasso algorithm \cite{arnold2007temporal}. Authors in \cite{dsouza2017exploring} propose a linear VAR to infer the topology in the original space by translating the VAR model in the linear principal space \cite{dsouza2017exploring}.  Authors in \cite{shen2017kernel} estimate a nonlinear function to assess the latent interactions of the time-series, to unravel the nonlinear dependencies, and in \cite{runge2019detecting}, the authors propose a conditional nonlinear independence hypothesis test in the Granger causality framework using Peter and Clark momentary conditional independence hypothesis tests ( PCMCI) \cite{runge2019detecting}.
Contemporary methods investigate linear time-lagged dependencies between the time-series vulnerable to confounding connections, see \textit{e.g.} \cite{marques2020signal,giannakis2018topology}; however, various methods have been proposed to infer the topology of networks with nonlinear interactions, see \textit{e.g.} \cite{shen2017kernel,runge2019detecting,giannakis2018topology,marques2020signal,wu2020discovering,marinazzo2008kernel,lim2015operator}.

The present paper extends the merits of the prior works, large-scale Granger causality \cite{dsouza2017exploring}, and puts forth a more general, nonlinear method, the lsKGC, for topology inference of directional time-series graphs. Unlike previous nonlinear efforts \cite{shen2017kernel, marinazzo2008kernel,wu2020discovering}, the present paper leverages the kernel principal component analysis as an encompassing framework to preserve the nonlinearity in low-dimensional space by projecting data into their principal components and utilizes pre-images of the feature space representations to infer the topology in the input space.  The present work estimates the model parameters in the low-dimensional feature space while estimates the topology of the graph in the linear framework of the input space.

%% file: simulations.tex
\vspace{-0.1in}
\section{Numerical Tests}
\begin{figure}
    \centering
    \includegraphics[width=0.45\textwidth]{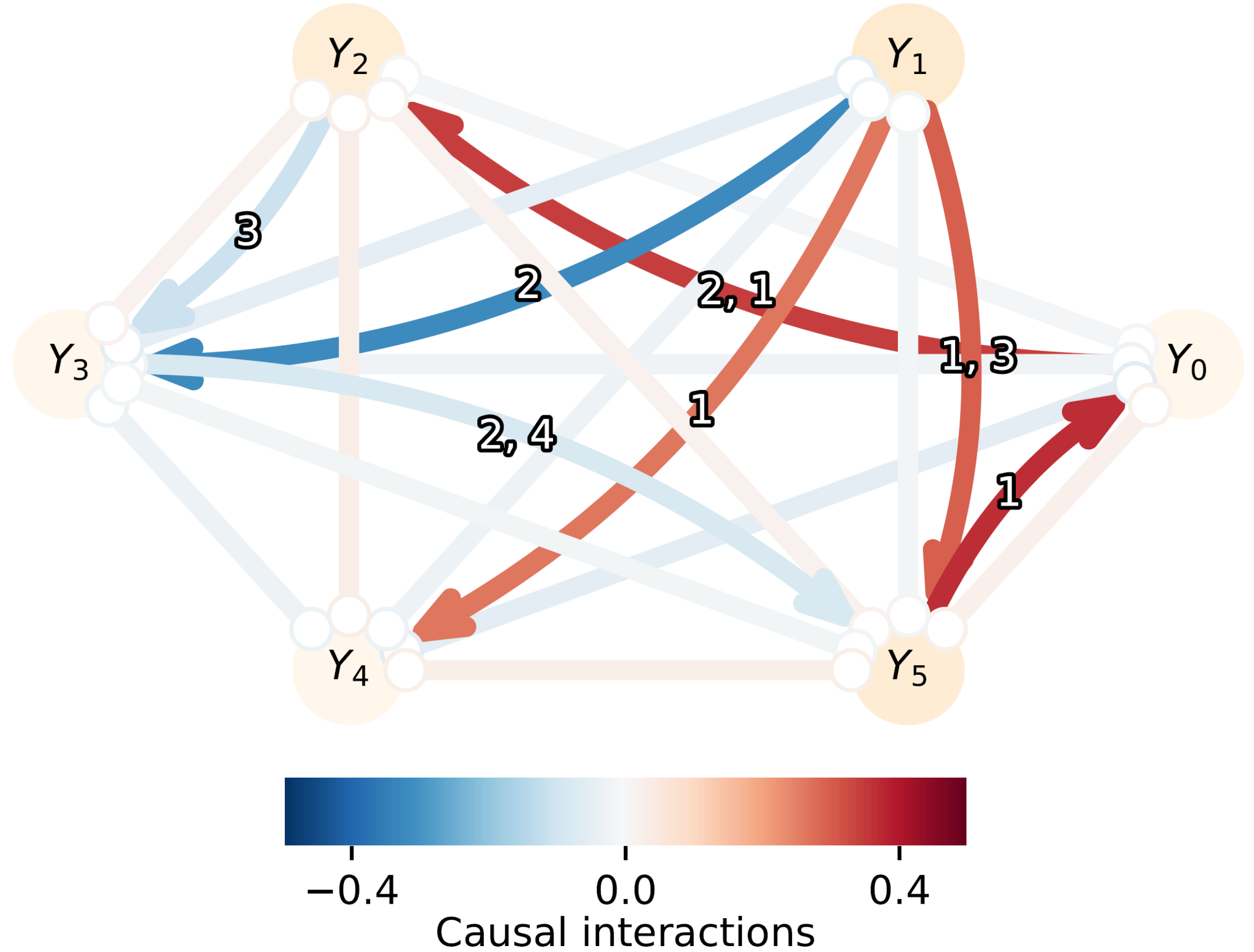}
     \caption{An instant graph of the underlying ground-truth network for the synthesized model with nonlinearities for N = 6. The color-bar represents the coupling's strength for the edges, and the number on the edges represents the lag.} \vspace{-0.25in}
     \label{fig:nonlinear_dependencies_time_evolve}
     
\end{figure}

\begin{figure}
\vspace{-0.3in}
\begin{subfigure}[b]{0.49\textwidth}
    \centering
    \includegraphics[width=\textwidth]{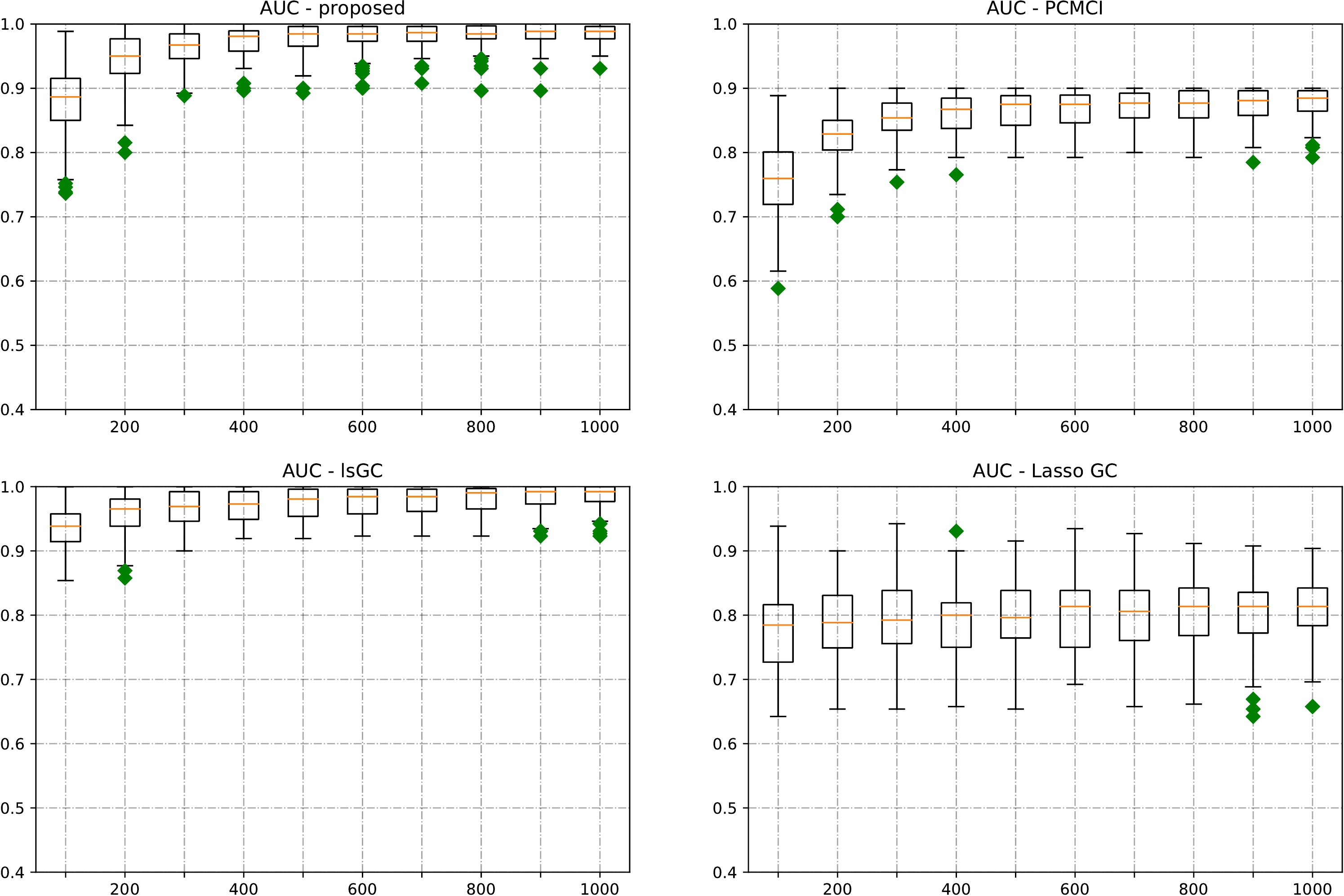}
    \caption{Topology inference tests on the synthetic datasets with only linear causal dependencies.}
    \label{fig:linear_with_lasso}
\end{subfigure}
\hfill
\begin{subfigure}[b]{0.49\textwidth}
    \centering
    \includegraphics[width=\textwidth]{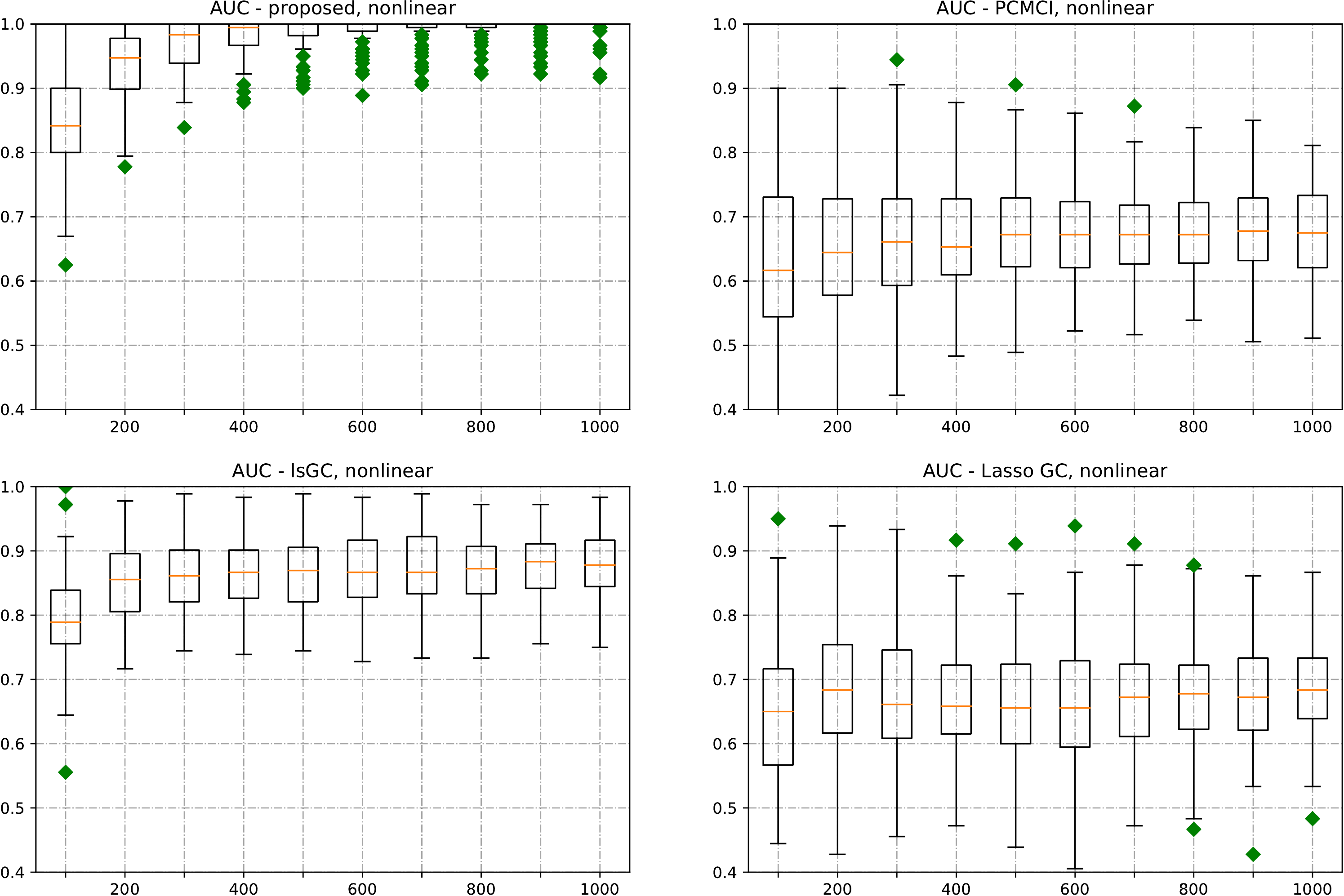}
    \caption{Topology inference tests on the synthetic datasets with both linear and nonlinear causal dependencies.}
    \label{fig:nonlinear_with_lasso}
\end{subfigure}
        \caption{Quantitative performance comparison for network topology inference on the synthetic datasets with linear and nonlinear causal dependencies. Box-plots of the proposed method as compare to PCMCI \cite{runge2019detecting}, lsGC \cite{dsouza2017exploring}, and Lasso-Granger \cite{arnold2007temporal} are shown. The vertical axis is the Area Under the Curve (AUC) for Receiver Operating Characteristics (ROC) analysis, where an AUC = 1 indicates a perfect network recovery and AUC = 0.5 random assignment. Whiskers are related to the 95\% confidence interval, green diamonds represent outliers, orange lines represent medians, and boxes are drawn from the first quartile to the third quartile. It is clearly seen that the proposed method outperforms PCMCI \cite{runge2019detecting}, lsGC \cite{dsouza2017exploring}, and Lasso-Granger \cite{arnold2007temporal} over all tested time-series lengths and topologies. It is clearly seen that the proposed method, along with , lsGC \cite{dsouza2017exploring}, outperforms PCMCI \cite{runge2019detecting}, and Lasso-Granger \cite{arnold2007temporal} over all tested time-series lengths and topologies.}
        \label{fig:three graphs}
\end{figure}

Simulations to evaluate and compare the proposed technique's performance include synthetic network datasets with known ground-truth and real fMRI datasets. The ensuing explains the description and results of tests on the datasets\footnote{The codes are available at https://github.com/ali-vosoughi/lsKGC}.
\newline\textbf{Synthetic dataset description.} To compare and evaluate the performance of the lsKGC, we construct synthetic datasets using \cite{runge2019detecting} with known underlying ground-truth. Generated datasets embody two kinds of dependencies: 1) Only time-lagged linear dependencies, and 2) including linear and nonlinear time-lagged dependencies between the time-series. 
Fig. \ref{fig:nonlinear_dependencies_time_evolve} shows a graph of the underlying network for N=6 of a synthetic network with nonlinear dependencies.  We generate several models with different random network topologies of T = 100,...,1000 time-series, with each network having linear or nonlinear causal dependencies. From each of these models, 100 time-series datasets are generated to assess the performance of each method.
\newline\textbf{Synthetic dataset results.} In an identical test setting, the boxplots in Fig. \ref{fig:linear_with_lasso} and Fig. \ref{fig:nonlinear_with_lasso} show a comparison of the large-scale kernelized Granger causality (lsKGC), PC-MCI \cite{runge2019detecting}, Lasso-Granger \cite{arnold2007temporal}, and lsGC methods \cite{dsouza2017exploring}. PCMCI (Peter and Clark momentary conditional independence) is a nonlinear independence test in the Granger causality framework for topology inference \cite{runge2019detecting}. Lasso-Granger is a method for topology learning based on the variable selection using Lasso \cite{arnold2007temporal}, and lsGC is a method for topology inference large-scale setting \cite{dsouza2017exploring}. The boxplots in Fig. \ref{fig:linear_with_lasso} and Fig. \ref{fig:nonlinear_with_lasso} show the  AUC distributions for each algorithm and various time-lengths. 
Numerical tests for models with linear dependencies in Fig. \ref{fig:linear_with_lasso} shows that the lsGC and the proposed method (lsKGC) competitively reach the top AUC (=1).  However, in tests for nonlinear models (Fig. \ref{fig:nonlinear_with_lasso}), the proposed method obtains a significantly higher AUC for most time lengths. 
\newline\textbf{Real dataset description.} The Centers of Biomedical Research Excellence (COBRE) data respiratory contains raw anatomical and functional MR data from 72 patients with schizophrenia and 74 healthy controls (ages ranging from 18 to 65 in each group) \cite{calhoun2012exploring}. All subjects of healthy controls and diseased patients under the age of 32 were selected, including 33 healthy and 29 diseased subjects, totaling 62 individuals. Functional connectivity measurements were generated from the COBRE dataset \cite{niak2016cobre}, a publicly available sample which we accessed through the Nilearn Python library \cite{nilearn}. The images were already preprocessed using the NIAK resting-state pipeline \cite{niak-preprocessed,niak2016cobre}. The number of regions of interest has been selected to be 122 with functional brain parcellations \cite{bellec2013mining}.\newline
\textbf{Real dataset results.} In the present study, brain connections served as features for classification and are estimated by the proposed large-scale kernelized method. Feature selection is performed on each training data set with k-fold cross-validation using \textit{Kendall’s Tau} rank correlation coefficient \cite{62_kendall1945treatment} and $10\%-90\%$ of test-to-train split ratio. For the 100 iteration cross-validation scheme for classification, the data set is divided into two groups: a training data set ($90\%$) and a test data set ($10\%$) in a way that the percentage of samples for each class was preserved. A Support Vector Machine (SVM) \cite{63_suykens1999least} is used for classification between healthy subjects and schizophrenia patients. All procedures are implemented in Python 3.8. We show 96.3\% (P=17, L=4) accuracy in the diagnosis of patients with schizophrenia (COBRE), while the same classification procedure using cross-correlation for the same dataset is 72.8\%.  The literature of classification accuracies on the COBRE dataset is listed in Table .\ref{tab:results}.  
\begin{table}[]\centering\caption{\label{tab:results} COBRE and ACPI datasets accuracies}
\vspace{-0.1in}
\begin{tabular}{|cc|cc|}
\hline
\multicolumn{4}{|c|}{\cellcolor[HTML]{96FFFB}COBRE dataset, Schizophrenia} \\ \hline
Paper             & Accuracy(\%)      & Paper           & Accuracy(\%)         \\ \hline
{\cite{rahim2017joint}}        & 86.2            & {\cite{plaschke2017integrity}}        & 61-72              \\ \hline
{\cite{wang2019multi}}       & 82.42           & {\cite{ramkiran2019resting}}        & 69                 \\ \hline
{\cite{xiang2020schizophrenia}}        & 93.1            & {\cite{cabral2016classifying}}      & 70.5               \\ \hline
{\cite{li2019machine}}        & 71.8            & This paper      & \textbf{96.28 $\pm$ 1.41}      \\ \hline

\end{tabular}
\vspace{-0.1in}
\end{table}